\documentclass[pmlr]{jmlr}
\usepackage{longtable}
\usepackage{booktabs}
\usepackage[load-configurations=version-1]{siunitx} 
 \usepackage{listings}
 
\definecolor{codegreen}{rgb}{0,0.6,0}
\definecolor{codegray}{rgb}{0.5,0.5,0.5}
\definecolor{codepurple}{rgb}{0.58,0,0.82}
\definecolor{backcolour}{rgb}{0.95,0.95,0.92}

 \lstdefinestyle{mystyle}{
    backgroundcolor=\color{backcolour},   
    commentstyle=\color{codegreen},
    keywordstyle=\color{magenta},
    numberstyle=\tiny\color{codegray},
    stringstyle=\color{codepurple},
    basicstyle=\ttfamily\footnotesize,
    breakatwhitespace=false,         
    breaklines=true,                 
    captionpos=b,                    
    keepspaces=true,                 
    numbers=left,                    
    numbersep=5pt,                  
    showspaces=false,                
    showstringspaces=false,
    showtabs=false,                  
    tabsize=2
}

\makeatletter
\def\set@curr@file#1{\def\@curr@file{#1}} 
\makeatother

\theorembodyfont{\upshape}
\theoremheaderfont{\scshape}
\theorempostheader{:}
\theoremsep{\newline}

\usepackage{bbm}
\usepackage{caption}
\usepackage{algorithm}
\usepackage[noend]{algpseudocode}
\usepackage{placeins}
\jmlrvolume{106}
\jmlryear{2020}
\jmlrworkshop{ Machine Learning for Healthcare}

\title[Clinical Documentation with Contextual Autocomplete]{Fast, Structured Clinical Documentation via Contextual Autocomplete}
 \author{\Name{Divya Gopinath} \Email{divyagop@mit.edu} \\
        \addr Department of Electrical Engineering \& Computer Science \\
       Massachusetts Institute of Technology\\
   Cambridge, MA, USA    \AND
       \Name{Monica Agrawal} \Email{magrawal@mit.edu} \\\addr Department of Electrical Engineering \& Computer Science \\
       Massachusetts Institute of Technology\\ 
       Cambridge, MA, USA
       \AND
       \Name{Luke Murray} \Email{lsmurray@mit.edu} \\ 
       \addr Department of Electrical Engineering \& Computer Science \\
       Massachusetts Institute of Technology\\
       Cambridge, MA, USA
       \AND
       \Name{Steven Horng} \Email{shorng@bidmc.harvard.edu} \\ 
       \addr Center for Healthcare Delivery Science \\
       Beth Israel Deaconess Medical Center\\
       Boston, MA, USA
       \AND
       \Name{David Karger} \Email{karger@mit.edu} \\ 
       \addr Department of Electrical Engineering \& Computer Science \\
       Massachusetts Institute of Technology\\
       Cambridge, MA, USA
       \AND
      \Name{David Sontag} \Email{dsontag@mit.edu} \\
      \addr Department of Electrical Engineering \& Computer Science\\
       Massachusetts Institute of Technology\\
       Cambridge, MA, USA
       }

\begin{document}

\maketitle
\begin{abstract}
  We present a system that uses a learned autocompletion mechanism to facilitate rapid creation of semi-structured clinical documentation. We dynamically suggest relevant clinical concepts as a doctor drafts a note by leveraging features from both unstructured and structured medical data. By constraining our architecture to shallow neural networks, we are able to make these suggestions in real time. Furthermore, as our algorithm is used to write a note, we can automatically annotate the documentation with clean labels of clinical concepts drawn from medical vocabularies, making notes more structured and readable for physicians, patients, and future algorithms. To our knowledge, this system is the only machine learning-based documentation utility for clinical notes deployed in a live hospital setting, and it reduces keystroke burden of clinical concepts by 67\% in real environments.
\end{abstract}

\section{Introduction}
Clinicians currently spend more time documenting information in electronic health records (EHRs) than communicating with patients, and the timesink in using inefficient EHRs is posited to be a leading cause of physician stress and burnout \citep{carayon15, gardner_burnout}. Doctors prefer using natural language and free-text for documentation over restrictive structured forms \citep{khorana}, but clinicians have adapted to time-intensive note-writing by relying on overloaded acronyms and jargon \citep{md_readability}. As an example, consider this sentence from a real Emergency Department (ED) clinical note: \texttt{pt w/ h/o MS}. While \texttt{MS} might represent \texttt{mitral stenosis} to a cardiologist, it also can be used to denote \texttt{multiple sclerosis} to other specialists. To a layperson, the clinical note may be incomprehensible unless acronyms are expanded: \texttt{patient with a history of multiple sclerosis}.

Consequently, medical documentation is often noisy, ambiguous, and incomplete. The lack of structure in notes further hinders understandability for patients, other physicians, and machines \citep{Aljabri2018, patient_engagement, Koch-Weser_language2009}. The information within EHR notes remains largely untapped and, at present, cannot be easily used for downstream medical care or for machine learning models that rely on structured data. 

\paragraph{Statement of Contributions} We propose a method called \emph{contextual autocomplete}, which quickly captures clinical concepts at the point-of-care via learned suggestions. To do so, we build a hierarchical language model for clinical concepts that can operate in the noisy domain of ED notes. Our model is designed to be deployed in a live hospital environment, with inputs constrained to the triage information and past medical notes available to a doctor \emph{before} a note is written. While these constraints make it infeasible to build a generative language model, we generalize to the task of \emph{autocompletion}, where we make multiple suggestions for the next clinical concept to document and allow the clinician to determine the correct choice. As all suggestions are mapped to standardized clinical vocabularies, we can simultaneously impose structure on notes as they are being written, disambiguate between concepts, and make documentation faster for clinicians in real-world hospital settings.

\begin{figure}
    \centering
    \includegraphics[height=4cm]{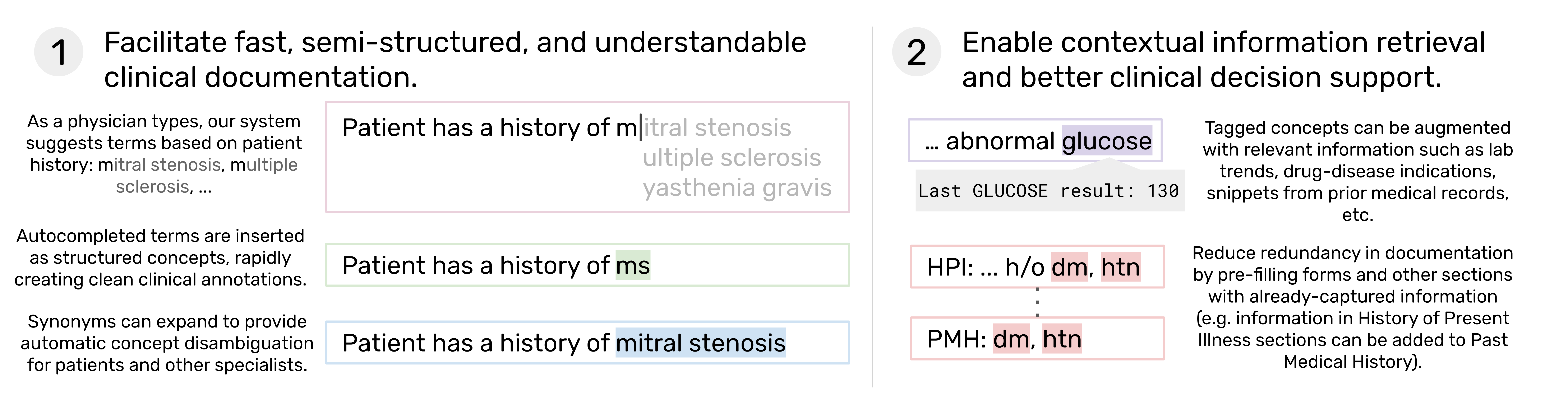}
    \caption{Semi-structuring notes with contextual autocompletion can enable extensive changes to both documentation and clinical decision support.}
    \label{typing_readability}
\end{figure}

\paragraph{Clinical Relevance} We present contextual autocompletion as the cornerstone of an intelligent EHR in Figure  \ref{typing_readability}. The contextual autocompletion tool reduces the amount of text a clinician has to type by suggesting relevant terms using a learned context. Tagged terms uniquely identify clinical concepts and can be linked with relevant information from the medical record. Moreover, these terms can facilitate widespread improvements in documentation and reduce overall cognitive load on doctors. Once a term is tagged, it can be automatically inserted in multiple locations within the clinical note to limit the amount of redundant information a clinician types-- for example, a tagged condition in an earlier part of a note can be automatically appended to the Past Medical History section that appears later on, as in the right panel of Figure \ref{typing_readability}. This mitigates the ``death by a thousand clicks" phenomenon that EHRs suffer from \citep{schulte_fry_2019}. In addition, live-tagging clinical concepts can provide immediate rewards to physicians in the form of decision support; the captured structured data can then be used to build smarter EHR interfaces that enable contextual information retrieval about disease history and lab trends, without ever leaving the note interface. Finally, tagging clinical concepts with our system allows for the translation of acronyms and domain-specific language to common names. By normalizing key clinical concepts from notes to a universal vocabulary (the Unified Medical Language System, or UMLS), notes written using our system are semi-structured and parseable.
\FloatBarrier
\paragraph{Generalizable Insights about Machine Learning in the Context of Healthcare} Prior research in clinical concept extraction tries to retrospectively recover concepts from notes, but these methods often struggle to disambiguate between similar concepts, and suffer due to a lack of labeled data \citep{mcn}. Even for a human expert, it is difficult to reliably disambiguate between concepts when the clinical intention is unclear. Building a tool that allows clinicians to document terms on-the-fly not only decreases documentation burden, but also curates large-scale prospective datasets of labeled clinical concepts (e.g. conditions, symptoms, labs, and medications) in notes. These labels can be used to design robust medical knowledge graphs, develop better clinical entity extraction models, learn longitudinal patterns within disease history, and even build contextual representations of concepts. Learning relationships between clinical concepts, for example, can inform and fill gaps in existing medical ontologies \citep{umls_missingness}. In addition, this work demonstrates the power of combining varying noisy and possibly incomplete sources of a patient's medical record to create a context for the current clinical setting that allows us to accurately predict concepts relevant to a new medical event. 
\FloatBarrier
\section{Background: The Current State of Clinical Workflows}
When a patient enters the ED, there are several phases of documentation. First, a triage nurse records patient vitals and a short description of the visit reason. This triage note is then summarized in a succinct phrase known as the Chief Complaint. Doctors also maintain a clinical note which is updated throughout the course of the visit and contains information about the patient's history, current presentation, pertinent labs and tests, and a final diagnosis and treatment plan. This note is also a constantly-evolving document. It is edited before the doctor sees the patient (to document patient history), while treating a patient (to document relevant symptoms and tests), and after the patient is discharged (to document the final diagnosis). The note is time-intensive to create, and as such, our work focuses on decreasing documentation burden within the doctor's note.

Clinical staff also have access to the patient's past EHR, which is a rich data source. The bulk of information in EHRs lies within unstructured clinical notes in the patient's file, which contain detailed information about disease history and prior clinical care. Yet these notes are long and difficult to quickly parse-- in our dataset, the median number of EHR notes per person is 34 with a median note length of 301 words. There have been attempts to mitigate this information overload by creating semi-structured representations of a patient's medical history such as the problem list, which catalogs a patient's prior conditions. However, these lists are poorly maintained and inconsistent amongst practitioners \citep{problem_lists}.

Efforts to intelligently structure free-text within clinical notes have been limited. One common technique is to pre-fill notes with templates that rely on structured text \citep{weis2014copy}-- for example, clinical notes usually begin with a summary of patient demographics and the chief complaint like \texttt{26 y/o M complains of dyspnea}. This method works for routine cases and structured, repetitive phrases that occur in some sections of notes, but fails to capture subtleties of documentation that reflect the nuances of clinical reasoning and physician preference.

To date, the largest-scale attempt to ease clinical documentation burden with machine learning is by \citet{GreenbaumChiefComplaint}, who built an autocomplete model to predict candidate chief complaints in the ED from a set of approximately 200 standardized options. The model --a multiclass SVM trained on triage information-- was used to structure 99\% of chief complaints in a live setting. We build on the work in \citeauthor{GreenbaumChiefComplaint} to provide contextual autocompletion functionality for an unstructured clinical note by architecting a model that incorporates both contemporaneous clinical information (triage text, vital signs, and laboratory results) and past medical history (EHR), and by building an interface that supports intuitive and on-the-fly documentation of multiple tagged terms from a large set of clinical concepts.
\FloatBarrier
\section{Methods}
\begin{figure}
    \centering
    \includegraphics[width=\textwidth]{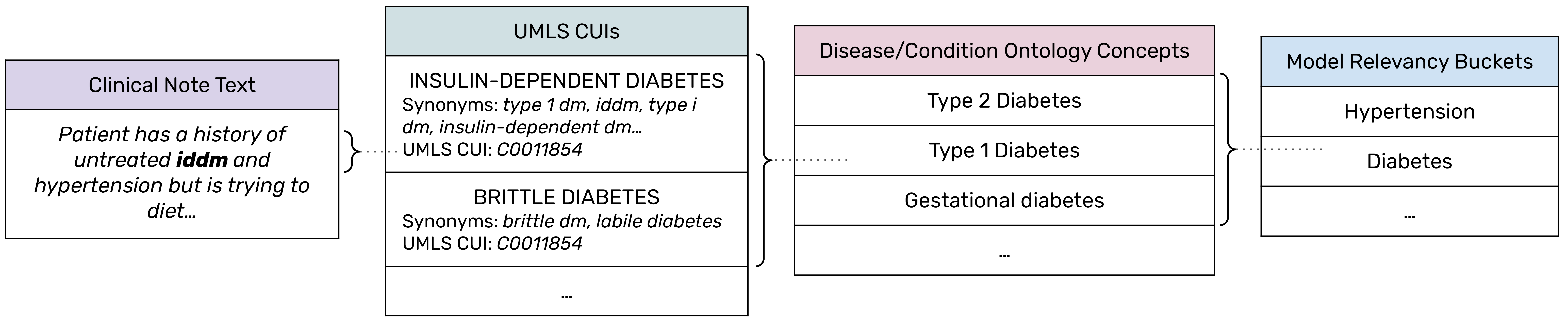}
    \caption{Data model for our ontology of conditions. Clinical notes are normalized to UMLS, and sets of UMLS IDs (CUIs) are aggregated to create unique concepts in our ontology. Ontology entries are then grouped together in coarser model relevancy buckets.}
    \label{ontology_diagram}
\end{figure}
\subsection{Data Overview and Cohort Definitions}
We use data from 273,000 anonymized visits to the Beth Israel Deaconess Medical Center (BIDMC) ED over the last decade, representing around 140,000 unique patients. For each visit, we have access to patient demographics, triage information (triage assessment, vital signs, and chief complaint), clinical notes from both the doctor and nurse assigned to the patient, and medications currently prescribed to the patient. In addition, we have all prior EHR notes for patients who had previously been in the BIDMC system (74\% of visits). We do not restrict our analysis to any particular subset of these visits.

\subsection{Defining Labels for Autocompletion}\label{dataset_labels}
The goal of autocompletion is to predict terms that the doctor would type into a note given a clinical context. In order to create positive labels for this task, we must extract documented clinical concepts from medical notes through named entity recognition (NER) on the text. We then normalize these concepts to UMLS. An example of this is shown in Figure \ref{ontology_diagram}.

First, we restrict ourselves to a subset of the UMLS ontology and exclude terms that do not correspond to a concrete clinical concept (e.g. \texttt{Health Care Activity}). The filtered terms are then inserted into a trie data structure, which we use to identify all UMLS concepts in time linear in the note length. We also apply a modified NegEx-style negation detection algorithm as in \citet{negex} to identify and mark which of these extracted terms occurred within a negative context. We provide additional details on our implementation of negative context detection in Appendix \ref{negex_appendix}. After filtering out concepts that appear fewer than 50 times, we extract 8,678 remaining UMLS concepts from visit notes. We then group concepts into two categories: conditions that a patient might have a history of, and symptoms that occur in the present medical context. Ambiguous acronyms such as \texttt{MS} are resolved as follows: if the term is almost always used to represent a singular concept within the ED, we default to that CUI, and otherwise ignore it. Two clinicians then independently verified these lists.

Concept disambiguation between closely linked conditions is difficult. As an example, \texttt{hyperlipidemia} and \texttt{increased LDL} are distinct UMLS concepts that encode similar semantic meanings, whose differences are not clinically meaningful in the ED and may be unknowable to the physician. To improve sample efficiency, we share weights between similar conditions by introducing a manually-curated hierarchy of UMLS terms and rolling terms up to an appropriate level of specificity such that every combined term carried the same medical meaning. A subset of this ontology for conditions is shown in Figure \ref{ontology_diagram}. The complete revised ontology (consisting of 940 entries encompassing 8,451 unique UMLS CUIs) can be found in Section \ref{ontologies_appendix} of the Appendix.

Condition concepts also represent varying levels of granularity, which is necessary in clinical text \citep{granularity_tange}-- a doctor could use \texttt{depression}, \texttt{severe depression}, or \texttt{chronic depression} to describe a patient, but these are distinct entries in our ontology. Choosing between similar terms during documentation is currently a subjective practice that depends on the clinical scenario and user-specific preference. We address this by further rolling up our ontology into a coarser set of \emph{model relevancy buckets} which group terms corresponding to similar underlying medical concepts. We build our models to have predictive power at the level of relevance buckets, and later rank individual terms within a model relevancy bucket to suggest terms for a doctor to document. This injects a medical inductive bias that forces parameter sharing between similar concepts, thereby allowing us to leverage closely-related groups of rare conditions to learn a common predictor. A subset of the 227 model relevancy buckets can be seen in Figure \ref{ontology_diagram}. We find our UMLS-based extraction is equally effective on our text as out-of-the-box learned extraction models such as cTakes \citep{ctakes}, scispaCy \citep{scispacy}, and DistilBERT \citep{Sanh2019DistilBERTAD}, while being the only paradigm suitably fast for live deployment. We elaborate on this in Section \ref{umls_extraction_appendix} of the Appendix. 

\subsection{Developing Predictive Features for Autocompletion}
In a typical language model, one attempts to predict the distribution $p(w_i|c_i)$ of an unknown word $w_i$ using a \emph{context} $c_i$ which captures the semantic information necessary to make such a prediction. For a generative model, $c_i$ usually consists of a complex representation of $w_{1:i-1}$ (the words preceding $w_i$) and is often parameterized by a deep neural network. These representations are state-of-the-art for clinical language modelling \citep{ranganath_bert, clinicalnlp_deep, liu_google}. In our framework, complex inference techniques are too slow to surface live suggestions with low latency in a hospital setting, and we only seek to predict clinical concepts rather than the general language a clinician types.

All features we use as part of our context must be available \emph{before} a patient and physician interact, as this allows us to surface live suggestions as the clinician creates documentation. This limits the data we can incorporate into our autocompletion models to unstructured textual data from the EHR, which is our glimpse into the patient's medical history; and triage-time information such as vitals, chief complaint, and the triage assessment. 

\subsubsection{Featurizing Textual Data}\label{featurize_text}
Our greatest sources of knowledge about the patient prior to clinician interaction lies in prior EHR notes and the triage assessment. To featurize prior EHR documents, we run the NER and hierarchical roll-up algorithms from Section \ref{dataset_labels}. The result of this is a mapping from a clinical text $T$ to a set of UMLS-mapped clinical concepts mentioned in the text, as well as a coarser representation of the types of conditions incorporated into the note. To encode triage assessments, we simply use a standard term frequency-inverse document frequency (TF-IDF) encoder to capture a normalized bag-of-words representation of the text.

\subsubsection{Featurizing Triage Vitals}
Triage vitals are already structured as they represent information that is inherently quantitative, such as heart rate and blood pressure. We discuss specific strategies of further preprocessing triage vitals with each model use-case below. 

\subsection{Autocompletion Models}
We frame contextual autocompletion as a hierarchical, human-in-the-loop language model that suggests clinical concepts to document as a physician is typing. We leverage four pieces of data to form our context $c_i=[w_{1:i-1}, T, \mathcal{H},V]$; namely, the text so far, the triage assessment, past EHR notes, and the patient's triage vitals. Calling an inference step of our model each time a word is written or removed is prohibitive in terms of latency, so we employ a rules-based approach to incorporate $w_{1:i-1}$ into our prediction while learning how to use the nuanced information in $T, \mathcal{H},V$. Inference thus only needs to be run once per patient. We first use $w_{1:i-1}$ to determine when the clinician wants to enter a potential clinical concept, and if so, whether that concept is a condition, symptom, lab, or medication. We then generate four term-wise rankings for each concept type, and stack the suggested rankings for each of the concept types to generate a total ranking. In practice, we filter these rankings to entries with any synonyms that match the typed query a doctor has entered. The doctor can either continue to type or select a term, which is then inserted into the note as a tagged concept using the synonym that he/she intended-- as an example, typing \texttt{ht} might give \texttt{hypertension} as a suggestion because of its synonym \texttt{htn}, and if a doctor chooses to autocomplete, we insert \texttt{htn} to preserve intended note vocabulary. We outline our four concept-specific ranking models:
\begin{enumerate}
    \item Conditions: we learn a mapping from the triage text and the clinical concepts mined from the EHR to a ranked list of relevant prior conditions that the doctor might want to document. This autocompletion model is primarily used to write the History of Present Illness (HPI) sections of notes, where physicians note past medical history that is relevant to the current patient presentation. We find that vitals have little to no predictive power in this model.
    \item Symptoms: we learn a mapping from the triage text, chief complaint, and vitals to a ranked list of relevant symptoms that the patient currently presents with. We do not include information from the patient's past medical record in our predictions because a patient's current presentation is only loosely related to prior visits.
    \item Labs: we simply list labs by their recorded frequency in $\mathcal{H}$, rather than learning a mapping. The space of labs is much smaller than the space of symptoms or conditions, so we find that a frequency-based ranking is nearly optimal in practice.
    \item Medications: As with labs, we rank by frequency for the same reasons.
\end{enumerate}
\subsubsection{Autocompleting Conditions}
Documenting relevant patient history is often an arduous task for physicians in the ED. Doctors typically read a patient's triage assessment and then search through a patient's EHR on an ad-hoc basis to try and contextualize the current visit with the patient's background. In our dataset, there is a median of 65 distinct conditions mentioned in a patient's EHR, but on average, only 5 of these concepts are then documented in the ED clinical note. In addition, around a quarter of the patients in our dataset do not have any prior records on file; in these cases, doctors can guess relevant conditions to inquire about based on the triage text and chief complaint alone.

This leads to key model desiderata: first, we must be able to recover an intelligent ranking over concepts even in the absence of prior medical notes using triage information alone. Second, we seek to learn a single multilabel ranking over all possible model relevancy buckets in order to produce a globally calibrated model. Our model first learns a ranking over the coarse model relevancy buckets, and then recovers a ranking over individual condition concepts to mention in the note. We use a shallow, dual-branch neural network architecture to combine a context $c_i$ consisting of a TF-IDF representation of the triage text $T$ and a feature vector indicating the binary presence $\mathbbm{1}[b \in \mathcal{H}]$ of each model relevancy bucket $b$ in prior EHR notes $\mathcal{H}$. Each arm of the network is passed through a single dense layer with rectified linear unit activation, the two outputs of the dense layers concatenated, and then the combined embedding is passed into a final dense layer with sigmoid activation to provide a vector of estimates of relevancy for each bucket. We recover a term-wise ranking by sorting each term first by whether it appears in the EHR, then by the rank of its relevance bucket, and finally by its empirical frequency of occurring in the data to resolve ties. In this way, we create a single architecture that predicts $P(b|T,\mathcal{H})$, or the probability of $b$ being relevant given the triage information and prior history, for all $b$ simultaneously and thereby suggest conditions to document for patients both with and without a prior medical history. Training details for this architecture can be found in Section \ref{training_appendix}. We also compare against three baselines:
\begin{enumerate}
    \item \textit{One vs. Rest Logistic Regression on Triage Text}: We build a model based solely on $T$. For each model relevancy bucket $b$, we estimate the $P(b | T)$ via a logistic regression model trained on a TF-IDF representation of $T$ to predict if any term in $b$ was mentioned in the corresponding clinical note. We randomly select notes without any mention of $b$ to generate negative samples. To make a prediction for a given patient, we then rank relevance buckets $b$ by $P(b|T)$. To recover a term-wise ranking, we sort each term first by the rank of its corresponding relevance bucket and by its empirical frequency in clinical notes. 
    
    \item \textit{One vs. Rest Logistic Regression on Triage Text, EHR}: As above, we train a logistic regression model on $T$ for each model relevancy bucket $b$. However, when predicting $P(b|T)$, we restrict ourselves to train on samples where $b$ is mentioned in $\mathcal{H}$. That is, our model predicts the probability $P(b|T,\mathbbm{1}[b \in \mathcal{H}])$. We assume $P(b|T, 0) = \epsilon_b$ for a small but nonzero $\epsilon_b$, or that if if a term does not appear in a patient's EHR, it is unlikely that it will be documented in the present note. To recover $P(b|T)$, we multiply by an empirically computed prior probability $P(\mathbbm{1}[b \in \mathcal{H}])$ of each bucket being mentioned in the EHR. We recover a term-wise ranking using the same key as the previous method. The leak probability $\epsilon_b$ allows us to rank buckets that are not present in the EHR by their empirical probabilities alone, giving us predictive power for patients without any prior history. 
    
    \item \textit{Augmented One vs. Rest Logistic Regression on Triage Text, EHR}: We experiment with feature-engineering approaches to include signals from the EHR in our model covariates. In particular, we augment the feature space with a representation $D$ of how many days it has been since $b$ was mentioned in the EHR, and compute $P(b|T,D,\mathbbm{1}[b \in \mathcal{H}])$ via logistic regression. In order to force this input variable to conform to a normal distribution, we transform the delay times by assuming mentions follow a Poisson process and concluding that delay times should be exponentially distributed. We follow the same empirical reweighting and term-wise ranking procedure as in the previous model.
\end{enumerate}
\subsubsection{Autocompleting Present Symptoms}\label{symptom_autocomplete}
Based on discussions with clinicians as well as qualitative analyses within our slice of ED data, we find that the symptoms that a doctor asks a patient about and subsequently records are primarily rule-based. A chief complaint of dyspnea at triage-time, for example, might prompt the doctor to inquire about dyspnea (reaffirming that it is still a concern), chest pain, coughing, etc. Consequently, the models we develop for symptom autocompletion use only the chief complaint and triage vitals as covariates. We perform ablation tests with all of our models to confirm that adding in a bag-of-words representation of the triage text did not increase performance, and develop four schemes to map chief complaints and vitals to a ranking over symptoms: 
\begin{enumerate}
    \item \textit{Empirical Conditioning on Chief Complaint:} For a given chief complaint $c$, we empirically calculate $P(s | c)$ for each $s$ in the set of symptoms $S$, and rank each symptom by this probability. 
    \item \textit{Empirical Conditioning on Chief Complaint, Vital:} For a given chief complaint $c$ and a list of vitals $V$, we calculate the single vital $v \in V$ that is most abnormal. Abnormality is defined as the percentile deviation from the population median of the vital value. We then encode $v$ as a categorical variable $b(v)$ based on medical guidelines about the given vital (for example, heart rate vitals are placed into one of three buckets: \texttt{LOW HR}, \texttt{NORMAL HR}, and \texttt{HIGH HR}). Full details about the bucketization  procedure can be found in the Section \ref{triage_vital_appendix}. Finally, we empirically calculate $P(s | c, b(v))$ for each $s \in S$, and rank each symptom by this probability. 
    \item \textit{One vs. Rest Logistic Regression:} For each symptom $s \in S$, we train a logistic regression model mapping the chief complaint and vital values to whether $s$ appears in the ED note corresponding to that visit. Then, we rank the output probabilities for each symptom.
    \item \textit{One vs. Rest Naive Bayes:} For each symptom $s \in S$, we train a Naive Bayes classifier mapping the chief complaint and vital values to whether $s$ appears in the ED note corresponding to that visit. Then, we rank the output probabilities for each symptom.
\end{enumerate}
In practice, we find that the second scheme performs best and we use this for deployment. Comparative performance for these models is detailed in Section \ref{results}. 

\subsubsection{Autocompleting Labs and Medications}
Autocompleting labs and medications is different from symptoms and conditions in a few marked ways. A patient's medical record contains structured information about prior lab tests and values, as well as medications and their dosages prescribed in the past. This is in contrast to symptoms and conditions which are almost always referenced in unstructured notes or free text. Concept disambiguation is less pertinent because there are structured representations of labs and medications, and there are already semi-structured lists of labs and medications that exist in clinical records. The primary value-add for physicians to tag a mention of a lab/medication in a note is instead to \emph{enable immediate information retrieval}. Tagging \texttt{HCT}, for example, can prompt the visualization or insertion of a patient's hematocrit trend. We thus add lab and medication autocompletion to be thorough in our data collection, and use a frequency-based autocompletion for both data types.

\subsubsection{Determining Autocompletion Scope and Type}
There are two components to displaying autocompletion suggestions: (1) the \emph{scope} of autocompletion, which determines when a clinician wants to document a concept; and (2) the \emph{type} of autocompletion, which determines a ranking over whether the clinician wants to document a condition, symptom, lab, or medication. A potential approach to this problem is to build a sequential model predicting whether the next word typed will be a clinical concept and its corresponding type, but this requires significant client-side infrastructure to curb model latency-- Gmail's Smart Compose system, for example, which surfaces dynamic suggestions of words to type from a neural language model, is only made possible via custom hardware and extensive system infrastructure \citep{gmail_smart_compose}. We discuss this further in Section \ref{discussion}. To build a system which can run live, we instead adopt a rule-based approach. 

We first define a default concept-type ranking per note section. For example, in HPI, the majority of documented content pertains to historical conditions and some current symptoms/medications, so the default ordering is \texttt{CONDITION, SYMPTOM, MEDICATION, LAB}. In contrast, in a Physical Exam section, clinicians document symptoms more than chronic conditions, so the default ordering is \texttt{SYMPTOM, CONDITION, MEDICATION, LAB}. We then establish certain key phrases to act as autocompletion triggers if they are likely followed by a clinical concept. We curate a list of common trigger phrases (e.g. \texttt{presents with}, \texttt{history of}) and map them to the concept type that follows them-- \texttt{presents with} is mapped to \texttt{SYMPTOM}, and \texttt{history of} to \texttt{CONDITION}. Using these, we create a NegEx-inspired algorithm to predict both autocompletion type and scope by greedily matching triggers in the text \citep{negex}. A full algorithm sketch of this is included in the Section \ref{negex_appendix}, and screenshots of the scope and type prediction algorithms at work are shown in Figure \ref{autocomplete_sc}.

While we rely on autocompletion scope and type prediction algorithms to guess where the user will insert a tagged term and the types of these terms, we support fallback data capture methods for when our algorithms fail. We do so in two ways. First, a user can start an autocomplete scope with a manual trigger. In addition, if the user does not type the manual trigger, we use an Aho-Corasick keyword detection algorithm to efficiently map exact string matches in the text with clinical concepts to our ontology \citep{aho_corsaick}. Any matches are displayed as potential tags which doctors can manually confirm if desired. A screenshot depicting these backup data capture strategies can be seen in Figure \ref{retrospective_annotation_sc} of the Appendix. We analyze how often these mechanisms are exercised in practice below.  Our rule-based algorithms have an average end-to-end latency of $\approx 0.2$ milliseconds to make a prediction, which is well below the 100ms threshold for a response to feel instantaneous \cite{nielsen_1993}. In contrast, making an API call to a shallow convolutional neural network for scope and type prediction takes upwards of 250ms in the absence CPU throttling, network overload, etc. 

\section{Results}\label{results}
\begin{figure}
    \centering
    \includegraphics[width=15cm]{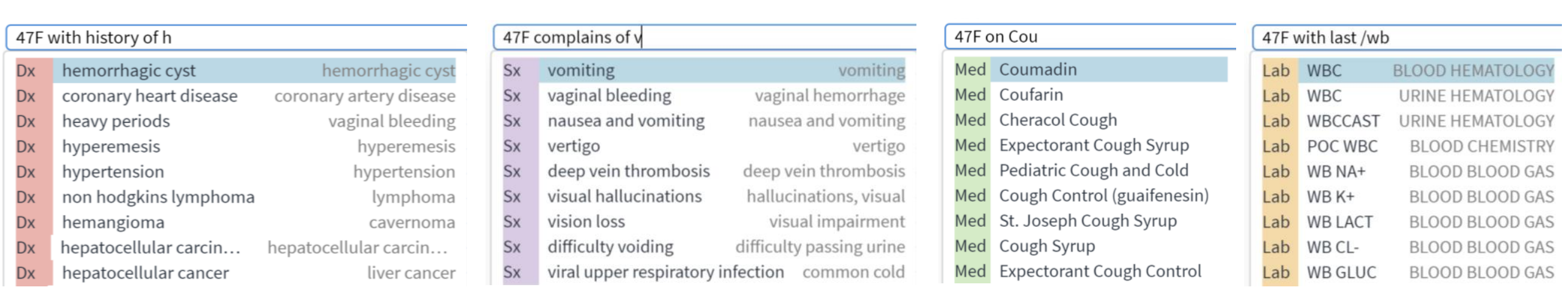}
    \caption{Screenshots of contextual autocompletion tool for each autocompletion type. From left to right: (a) Conditions (b) Symptoms (c) Medications and (d) Labs. Trigger words before the tagged term affect the scope and type of the autocompletion. Clinical concepts with synonyms that match the typed text are listed with the synonym in black text and the more general concept name in gray.}
    \label{autocomplete_sc}
\end{figure}
A physician uses contextual autocompletion by naturally typing a note and either automatically or retroactively completing clinical phrases that are then rendered as tagged concepts. As in standard autocomplete, autocompleted concepts are filtered to those that match the user's typed prefix. We briefly describe the user experience of the tool with a screenshot in Figure \ref{autocomplete_sc}, and examine how it reduces clinical documentation burden in practice.
\FloatBarrier
\subsection{Performance and Usability Metrics}
\subsubsection{Retrospective Evaluation on Clinical Notes}
Before deploying our autocompletion models in a live setting, we evaluated the quality of our suggested rankings via retrospective annotation of the clinical notes we had on file. In particular, we measured performance broken down by concept type, as well as the efficacy of our autocompletion scope and type detection algorithms. We use two standard information retrieval metrics, the \emph{mean reciprocal rank} (MRR) and \emph{mean average precision} (MAP) to gauge the quality of our rankings (definitions in Section \ref{retrospective_performance_appendix}). We compare the MRR of our contextual autocompletion tool rankings against two naive baselines: spell-based autocompletion (ranking terms alphabetically) and frequency-based autocompletion (ranking terms by frequency). Performance by MAP is detailed in the Section \ref{retrospective_performance_appendix}.

To generate our evaluation set, we extract medical concepts from 25,000 clinical notes with the technique outlined in Section \ref{featurize_text}. Using the order in which concepts were suggested, we first measure MRR assuming our scope and type detection was perfect, broken down by the four concept types. Results are shown in Figure \ref{retrospective_evaluation}. We see the largest gain in using a contextual model for conditions, because the space of terms is large and the richness of the EHR greatly influences documentation. Within the contextual models for predicting prior conditions, the dual-branched neural network outperforms others primarily because it is predictive even for patients who did not have any history on file at the hospital. On the other hand, when documenting symptoms, a model that ranks symptoms by their empirical frequency (conditioning on the chief complaint and the most abnormal vital) performs best.

To quantify the ease of documentation using our autocompletion scope and type detection algorithm, we also measure MRR when typing HPI sections of notes. We focus on HPI notes as they contain a range of concept types (conditions, symptoms, medications, etc.) and also were the only note section we could reliably segment due to dataset limitations. On average, there are 6.8 documented clinical concepts per HPI section. Of the extracted clinical concepts in HPI sections, 46\% of terms were autocompleted automatically without a manual trigger, and in 77\% of those cases, we guessed autocompletion type correctly as well. As a result, the MRR of automatically-detected autocompleted terms is 0.35. Even in cases where the doctor is forced to insert a manual trigger to autocomplete a term, we still greatly decrease the documentation burden on doctors as shown in Figure \ref{retrospective_evaluation}. These manually-prompted scenarios can be mitigated as a doctor learns and adapts to the autocompletion triggers of the system, which we elaborate on in Section \ref{discussion}.

\begin{figure}[t]
    \begin{center}
    \small
    \subfigure[Comparison of MRR between contextual autocompletion models]{
    \begin{tabular}{p{8cm}|p{1.5cm}}
    \toprule 
    Model Type & MRR $\uparrow$ \\
    \midrule 
     \textbf{Conditions} & \\ 
     One vs. Rest Logistic Regression on $T$ & 0.09 {\tiny $\pm 0.02$} \\ 
     OvR LR on $T$, EHR & 0.15 {\tiny $\pm 0.02$} \\
     Augmented OvR LR on $T$, EHR & 0.17 {\tiny $\pm 0.01$} \\
     Dual-branched neural network & \textbf{0.28} {\tiny $\pm 0.01$} \\ 
     \midrule 
     \textbf{Symptoms} &  \\ 
     Empirical Conditioning on Chief Complaint &  0.39 {\tiny $\pm 0.01$}\\
     Empirical Conditioning on Chief Complaint, Vital & \textbf{0.42} {\tiny $\pm 0.01$} \\ 
     One vs. Rest Logistic Regression & 0.16 {\tiny $\pm 0.01$} \\ 
     One vs. Rest Naive Bayes & 0.27 {\tiny $\pm 0.02$}\\
     \bottomrule
    \end{tabular}
    }
    \quad
     \subfigure[Comparison of MRR across autocomplete types]{
     \begin{tabular}{p{3.8cm}|p{1.6cm}p{1.6cm}p{1.6cm}}
    \toprule 
    Model Type & \multicolumn{3}{c}{Autocomplete Type} \\
    \midrule 
     & Spell & Frequency & Contextual \\ 
     \midrule 
     \textbf{Conditions} &  0.01 {\tiny $\pm 0.001$} & 0.08 {\tiny $\pm 0.01$} & \textbf{0.28} {\tiny $\pm 0.01$}  \\ 
     \midrule 
     \textbf{Symptoms} & 0.05 {\tiny $\pm 0.001$}\ & 0.27 {\tiny $\pm 0.01$}  &  \textbf{0.42} {\tiny $\pm 0.01$} \\
     \midrule
     \textbf{Labs} & 0.01 {\tiny $\pm 0.001$}& 0.40 {\tiny $\pm 0.01$}&  N/A \\
     \midrule
     \textbf{Medications} & 0.02 {\tiny $\pm 0.001$}&  0.02 {\tiny $\pm 0.001$} & N/A \\ 
     \midrule
     \textbf{Overall} & 0.01 {\tiny $\pm 0.001$} & 0.19 {\tiny $\pm 0.03$} & \textbf{0.29} {\tiny $\pm 0.05$}\\
     \bottomrule
    \end{tabular}
    }
    
    \caption{Retrospective Evaluation of MRR using Contextual Autocompletion. We report average MRR ($\pm$95\% confidence interval of the mean) for each of our learned contextual autocomplete models, and compare our best models (dual-branched neural network for conditions, empirical conditioning on the chief complaint and most abnormal vital for symptoms) to spell-based and frequency-based baselines, both for specific concept types as well as overall using our scope and type prediction algorithms. Calculated across 25,000 visits.}
    \label{retrospective_evaluation}
    \end{center}
\end{figure}
\FloatBarrier
\subsubsection{Documentation in the Wild: Live Evaluation}
Because the primary goal of this tool is to improve documentation efficiency, we also define the \emph{keystroke burden} as the number of keystrokes the clinician needs to type until he/she autocompletes and inserts a desired term. This usability metric inherently encompasses the quality of our information retrieval in its calculation while also incorporating real-world behavior-- there may be a delay between a term being suggested first and when a clinician actually autocompletes the term. We compare keystroke burden between a contextual model and no autocompletion in Figure \ref{live_evaluation}. In our live evaluation, a single physician wrote 40 notes using our system over two shifts. In practice, an average of  8.38 terms are tagged per note, and we reduce overall keystroke burden for these clinical concepts by approximately 67\%, with clear gains in using our model irrespective of note section or concept type. 53\% of the tagged clinical concepts were autocompleted without a retroactive label. Moreover, 96\% of these terms were automatically prompted (as opposed to the user manunally prompting the autocomplete), indicating our scope and type detection had high recall even when doctors had not yet adapted to the system. For 77\% of the terms tagged without a retroactive label, we also predicted the clinical concept type correctly. 

\begin{figure}[]
    \centering
    \small
    \begin{tabular}{p{5cm}|p{2cm}p{2cm}}
    \toprule 
    Subset & \multicolumn{2}{c}{Autocompletion Type} \\
    \toprule
     & None & Contextual \\ 
     \midrule
     Overall & 11.85 {\tiny $\pm 1.94$} & 4.32 {\tiny $\pm 0.43$} \\
     \midrule 
     \textbf{By Note Section} & & \\
     History of Present Illness & 12.36 {\tiny $\pm 2.16$} & 4.57 {\tiny $\pm 0.87$} \\
     Past Medical History & 11.41 {\tiny $\pm 2.09$} & 2.94 {\tiny $\pm 0.68$} \\
     Medical Decision Making & 10.27 {\tiny $\pm 3.18$} & 4.08 {\tiny $\pm 0.49$} \\
     \midrule 
     \textbf{By Concept Type} & & \\
     Conditions & 13.08 {\tiny $\pm 1.72$} & 4.34 {\tiny $\pm 1.49$} \\
     Symptoms & 8.5 {\tiny $\pm 2.18$} & 4.53 {\tiny $\pm 1.00$} \\
     Labs & 10.33 {\tiny $\pm 5.76$} & 2.06 {\tiny $\pm 0.88$} \\
     Medications & 9.27 {\tiny $\pm 1.97$} & 4.27 {\tiny $\pm 1.34$} \\
     \bottomrule
    \end{tabular}
    \caption{Live Evaluation of Contextual Autocompletion Models. Mean keystroke burden for autocompleted concepts ($\pm$ 95\% CI from mean), measured across 40 notes written live by a single physician over two shifts. Performance is also broken down by note section, as well as concept type. }
    \label{live_evaluation}
\end{figure}

\subsection{Autocompletion Sensitivity Analysis}
Concept frequency influences the efficacy of our contextual autocompletion model of conditions. The biggest wins in the model occur with the group of conditions in the middle of the frequency distribution-- \texttt{renal insufficiency}, for example, is an infrequent but not rare term that will almost certainly be documented in a note if it appears in the patient's history. The symptom contextual autocompletion model, on the other hand, is generally agnostic to concept frequency because the space of symptoms is much smaller and the distribution of symptoms is less skewed than that of conditions.

In addition, the presence of prior medical history has significant impact on contextual autocompletion performance for conditions-- as shown in Figure \ref{EHR_info_analysis}, we see greater reduction in documentation burden if the patient has prior EHR. However, our contextual model and a frequency-based autocompletion model perform similarly for concepts that are not mentioned in the EHR despite the person having some prior medical history-- this can largely be attributed to the inherent bias of our ranking scheme, which preferentially orders terms mentioned in the EHR above those that are not. 
\begin{figure}[]
    \centering
    \small
    \begin{tabular}{p{6.3cm}p{2cm}p{1.8cm}p{2cm}}
    \toprule 
    \multicolumn{4}{c}{\textbf{Mean Keystrokes Saved per Condition Concept}} \\
    \midrule 
     & Uncommon Concepts & Median Concepts & Common Concepts \\
         \midrule 
    With no past EHR at hospital & 0.63 {\tiny $\pm 0.42$} & 0.81 {\tiny $\pm 0.50$}& 0.47 {\tiny $\pm 0.20$}\\ 
    With prior mention of concept in EHR  & 2.64 {\tiny $\pm 0.65$} & 2.02 {\tiny $\pm 0.38$} & 1.40 {\tiny $\pm 0.16$}\\
    \bottomrule
    \end{tabular}
    \caption{Number of keystrokes saved by our contextual model compared to a frequency-based baseline ($\pm$95\% CI of the mean) for conditions. Performance was stratified by concept frequency (by terciles) and by available medical history.} 
    \label{EHR_info_analysis}
\end{figure}
\FloatBarrier
\subsection{Interpreting Contextual Autocompletion of Prior Conditions}
Because our contextual model for conditions learns a ranking from a representation of the triage text and medical history, it is naturally more sensitive to changes in input than our contextual model for symptoms. Here, we dig further into what drives model predictions. 

\subsubsection{Performance By Concept}
\begin{figure}[]
    \small
\begin{tabular}{p{2.8cm}p{5.5cm}p{6cm}}
    \toprule
     \textbf{Concept} & \textbf{Most Predictive Triage Tokens} & \textbf{Most Predictive Model Relevancy Buckets}\\
     \midrule 
     \textbf{Dementia} &  
     dementia, abrasions, fell, home, fall, neuro, son, ...& 
     dementia, neurodegenerative diseases \\
     \midrule 
     \textbf{Bronchitis} & 
     pna, pneumonia, cough, sob, hemoptysis, sputum, ... & 
     pneumonia, chronic lung disease \\
     \midrule 
     \textbf{Prostate cancer} &  
     ca, mass, chemo, lymphoma, melanoma, cll, tumor, .. & 
     cancers, prostatectomy \\ 
     \midrule 
     \textbf{CHF} & 
     chf, chest, sob, cp, cough, syncope, fall, ... & heart failure, heart attacks, hypertension, afib  \\ 
     \midrule
     \textbf{Diabetes} &
     bs, fsbs, glucose, iddm, sugars, toe, finger, ...  & diabetes, hyperlipidemia, diabetic neuropathies, gastroparesis \\ 
     \bottomrule 
\end{tabular}
    \caption{Predictive features for selected condition concepts, using a linear approximation to our contextual model for conditions. Inputs to the model are a TF-IDF representation of the triage text as well as the presence of coarse-grained model relevancy buckets in a patient's prior medical record, as defined in Section \ref{dataset_labels}.}
    \label{global_interpretability}
\end{figure}
Our multi-label model predicts the binary relevance of each model relevancy bucket. To better interpret relevancy predictions on a per-bucket level, we approximate our model for a specific relevancy bucket $b$ with a linear function of the inputs. This is done by fitting a $L_1$-regularized linear approximation between the features and the logits generated by the model for bucket $b$ to surface highly-weighted features \citep{knowledge_distillation}. In Figure \ref{global_interpretability}, we provide examples of the top-weighted positive features in the linear approximations to models for five selected concepts. Overall ranking performance by MRR for these concepts is in Figure \ref{mrr_by_concept} of the Appendix. Interestingly, while all of the chosen concepts relied on medically meaningful tokens present in the triage text, the linear models for diabetes and congestive heart failure both used the presence of many model relevancy buckets, whereas the other three concepts only relied on a few. This is likely because the model always relies on triage text but can give predictions even in the absence of prior medical history, and as the linear approximation to our model encourages sparsity, only highly predictive model relevancy buckets will be chosen as features. A frequency-based baseline outperforms our learned model only for extremely common conditions like hypertension and diabetes. 

\subsubsection{Qualitative Evaluation \& Readability} 
\begin{figure}[]
    \centering
    {
    \subfigure[Effect of patient history on contextual rankings.]{
     \includegraphics[height=4.2cm]{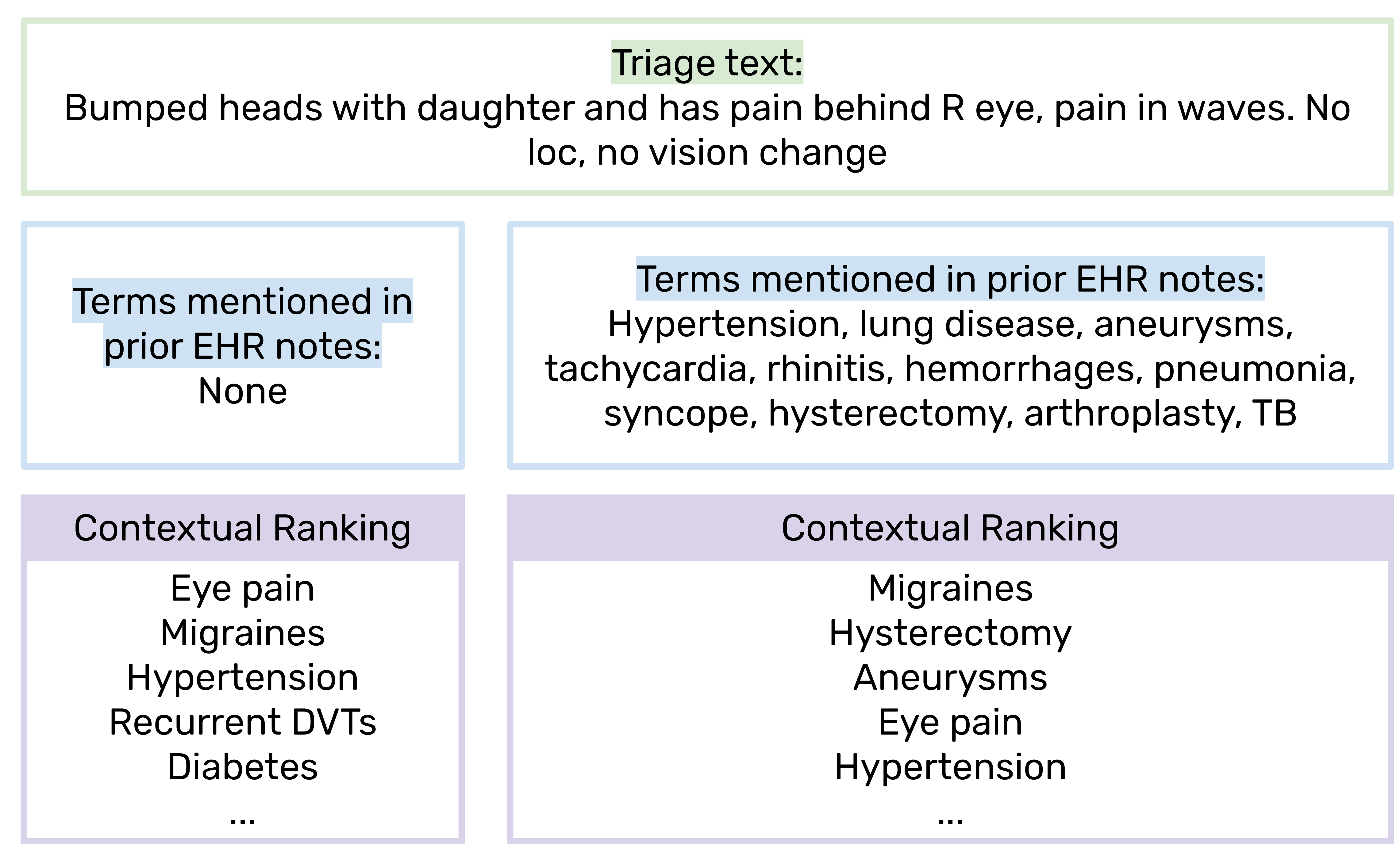}
     }
     \subfigure[Effect of triage note on contextual rankings.]{
     \includegraphics[height=4.2cm]{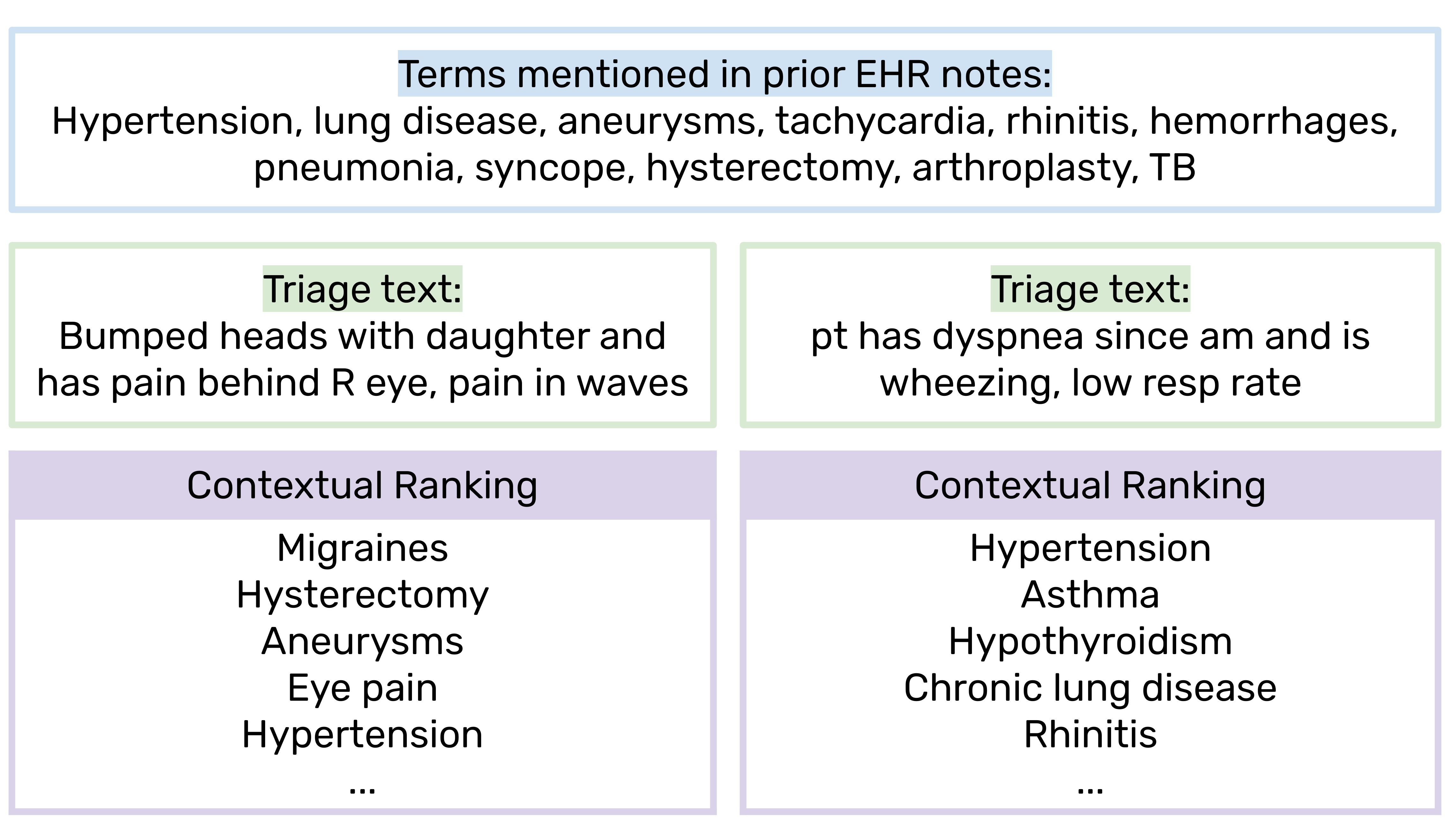}
     }
    }
    \caption{Case Studies of Rankings over Conditions}
    \label{case_studies}
\end{figure}

We qualitatively evaluate rankings over conditions to better understand model decisions. As can be seen in the selected examples in Figure \ref{case_studies}, both the presence of EHR notes as well as specific types of words mentioned in the triage note can have great impact on the rankings, which are much more context-specific than frequency-based rankings. Chronic conditions mentioned in a patient's medical history are highly ranked even if they are not directly related to the present medical context, because they are likely to be documented regardless. For example, in Figure \ref{case_studies}a, two patients have identical triage text but different medical histories-- consequently, \texttt{hysterectomy} is highly ranked for one. Of course, the triage note still governs the overall theme of the most highly ranked terms; in Figure \ref{case_studies}b, two patients with identical medical histories but differing chief complaints have vastly different context-specific rankings.
\FloatBarrier
\section{Discussion} \label{discussion}

The contextual autocompletion tool we have outlined harnesses the power of machine learning to encode information about medical contexts, and then uses this to suggest terms to document to clinicians. Medical professionals who utilize this tool can not only document terms more easily and save valuable time to interact directly with patients, but also can create clean annotations of clinical text in a novel manner. These annotations can be used to provide disambiguation between overloaded terms, clarify associations between medical concepts, and generate large-scale EHR datasets for future innovation. All of the medical ontologies built for this work map to UMLS, making our contextual autocompletion tool translatable to other clinical centers with minimal modification. The ablation tests we carried out show that using a few features (primarily representations of medical histories) can result in performant predictive models for documentation. This is a critical advantage of our system because EHR data is often very sparse-- patients can enter the ED with no prior medical history, yet we can still glean information from the triage assessment to represent a patient state. Our strategies also obviate the need for complex data imputation schemes. 

Our system provides automatic and natural autocompletion of a clinical concept when our scope detection algorithm is accurate, and users may need to resort to a manual trigger in other cases. This is not a major hindrance based on our evaluation criteria, and we significantly reduce documentation burden even if the manual trigger is used. However, we note that more complex model classes (e.g. recurrent neural networks/sequential learning models) may be better at scope prediction. These models can introduce significant client-side latency because they require performing inference after each character-level change of the note. Thus, to allow for straightforward integration into the existing hospital ecosystem, we do not explore these schemes in this paper. Future iterations of our tool might automatically learn a set of autocompletion triggers rather than using hand-crafted rules. Using manual triggers for autocompletion, however, can also establish consistent system behavior for physicians, and create notes that are concise. As an example, one clinical note in our dataset began with the phrase \texttt{patient has a history of abdominal pain which seems recurrent}, whereas our system would autocomplete to \texttt{patient has a history of \emph{chronic abdominal pain}}.

We propose two other future directions to build on and continue this work. The first is to better integrate key semantic modifiers into our tool. As an example, doctors often document the absence of symptoms (e.g. \texttt{no fever}) to aid in a differential diagnosis. While we can use rule-based approaches to retrospectively attach negation modifiers to tagged medical concepts, future work should seek to fuse modifier capture with the UI. Clinicians might also type a term that refers to someone other than the patient (e.g. \texttt{family history of diabetes in mother}), and we should automatically learn that the concept \texttt{diabetes} refers to a third-party rather than the patient.

In addition to facilitating semantic modifier capture, a next iteration should dynamically update suggested terms to document using already-tagged terms in the note. Tagging \texttt{atrial fibrillation}, for example, might indicate that there is a high likelihood of the doctor typing an anticoagulant next. Using live data collected from the deployed tool, we can use early drafts of a clinical note to influence the medical context for later autocompletion suggestions. We can also clarify patterns of redundant data entry by examining where the same underlying medical concept is repeated in the note, with the eventual goal of learning and auto-inserting necessary repetitious documentation. These dynamic updates introduce a significant latency on the client-side UI to perform online inference as words are typed, so this may not be feasible for all systems and thus we did not consider it in this first iteration.

\section{Conclusion}
EHRs have introduced significant burden on physicians, and to adapt, doctors have resorted to using overloaded jargon that then renders clinical notes unusable for downstream clinical care. The lack of clean labels for unstructured text also inhibits how we can utilize machine learning techniques to transform healthcare. There is a real need to modernize and exploit the information hidden within notes without interrupting the clinical workflow.  While our contextual autocompletion tool can reduce documentation burden and curate clean data for machine learning purposes, it also opens the possibility of reforming clinical documentation practices to make notes more understandable to humans and algorithms alike. Fundamentally, live-tagging of medical concepts enables unprecedented changes to EHR design. By integrating machine learning methodologies into documentation practices, we can usher in a new era of EHRs that assist rather than impede physicians. 

\acks{We thank the MIT Abdul Latif Jameel Clinic for Machine Learning in Health for the grant that funded this work.}

\bibliography{ref}
\newpage
\appendix
\section{Data Extraction and Featurization}
\subsection{Examples of Clinical Notes}
Here, we show examples of a triage note, chief complaint, patient vitals, and a clinician note. To preserve patient privacy, these examples are synthetic but mimic the formatting and style of real data.

\textit{Triage Note}
\begin{verbatim}
    pt with ruq abd pain and nonproductive cough
\end{verbatim}

\textit{Chief Complaint}
\begin{verbatim}
    ruq abd pain
\end{verbatim}

\textit{Vitals}
\begin{verbatim}
    Blood Pressure: 140/90 mmHg
    Heart Rate: 109 BPM
    Pain: 8 (out of 10)
    Sex: F
    Age: 66
    Respiratory Rate: 92%
    Temperature: 99 (deg. Fahrenheit) 
    Pulse Oxygen (Oxygen Saturation): 96
\end{verbatim}

\textit{Clinical Note} 
\begin{verbatim}
    HPI: 66 y/o F p/w ruq abd pain and nonproductive cough.
    No fever, nausea, or chills.
    History of chronic abdominal pain over last 4-5 years,
    as well as htn and dmii. 
    
    PMH: htn, dmii, chronic abdominal pain, hysterectomy in 2004
    
    MEDICATIONS: metoprolol tartrate, metformin
    
    FAMILY HISTORY: Diabetes in mother,
    father (deceased) hypertensive
    
    SOCIAL HISTORY: no smoking, drinks socially 
    
    REVIEW OF SYSTEMS: 
    Constitutional - no fever, chills, nausea
    Head / Eyes - no diplopia
    ENT - no earache
    Resp - nonproductive cough, mild
    Cards - no chest pain
    Abd - ruq abd pain
    Flank - no dysuria
    Skin - no rash 
    Ext - no back pain
    Neuro - no headache
    Psych - no depression
    
    PHYSICAL EXAM: Ruq abd pain, tender to touch, 
    with some bloating.
    
    MDM:
    66 y/o F p/w ruq abd pain and mild cough. She reports
    she had a cold last week, so cough
    is likely symptom of that. 
    
    Epigastric pain with mild bloating and minor
    heartburn. Gave an antacid to relieve pain. 
    
    Glucose levels are elevated compared to baseline 
    (140 6 hours ago, 120 averaged over last six months).
    Says she will work on controlling diet more. 
    
    DIAGNOSIS: epigastric pain/heartburn
    
\end{verbatim}

\subsection{The NegEx Algorithm}\label{negex_appendix}
We use a version of the NegEx algorithm \cite{negex} in order to perform a rule-based negation detection on clinical text. The algorithm greedily iterates through words in a piece of text and assigns them to a negated context if they are preceded by predefined keyword triggers. Pseudocode for the algorithm is shown in Figure \ref{negex_pseudocode}.

\begin{figure}
\begin{lstlisting}[language=Python]
fullstops = ['.', '-', ';']
midstops = ['+', 'but', 'and', 'pt', '.', ';', 'except', 'reports', 'alert', 'complains', 'has', 'states', 'secondary', 'per', 'did', 'other', 'p/w', 'presents', 'presenting', 'presented', ':']
negwords = ['no', 'not', 'denies', 'without', 'non', 'lack']

def negation_detection(words):
    flag = 0
    res = []
    for i, w in enumerate(words):
        neg_start_condition = (flag == 1)
        neg_stop_condition =  (w in fullstops + midstops + negwords) or (i > 0 and words[i-1][-1] in (fullstops + ['\n']))
        neg_end_of_list = (i==(len(words)-1) )
        if neg_start_condition and neg_stop_condition:
            flag = 0
            res += [(start_index, i-1)]
        elif neg_start_condition and neg_end_of_list:
            flag = 0
            res += [(start_index, i)]
        if w in negwords:
            flag = 1
            start_index = i
    return res
\end{lstlisting}
    \caption{Pseudocode of the rule-based negation detection algorithm.}
    \label{negex_pseudocode}
\end{figure}

\subsection{Trie-Based Extraction of UMLS Concepts}\label{umls_extraction_appendix}
In order to confirm that our UMLS-mapped trie-based extraction of clinical concepts was reasonably accurate and performant, we also consider a few alternate ways of perform clinical NER on ED note text. We restrict our search to techniques that normalize to UMLS, as this is a key benefit of our system that makes it extendable.

First, we attempted to extract concepts directly from the raw text, without normalizing to an ontology. We did this by extracting common unigrams and bigrams and removing common stopwords (\texttt{and}, \texttt{to}). We manually went through the 1,000 most common terms to confirm they were reflected in our UMLS-mapped ontology of conditions, and added a handful of terms that were missing: \texttt{hld} as a synonym for hyperlipidemia, \texttt{hep c} as a synonym for hepatitic C, \texttt{pna} for pneumonia, etc. We note that ontologies are always a work in progress and that our current system provides doctors with the ability to submit ontology modifications that can then be reviewed. 

We compare our trie-based extraction against three baselines: 
\begin{itemize}
    \item \texttt{cTakes}, or the Mayo clinical Text Analysis and Knowledge Extraction System, which combines rule-based and simple machine learning techniques to extract and normalize concepts to UMLS \citep{ctakes}. cTakes is an older system that often misses clinical abbreviations \citep{Rojas2018ComparisonOM}. We limit the cTakes vocabulary to UMLS concepts in our ontology to provide a fair comparison. 
    \item \texttt{scispaCy}, which is a Python biomedical text processing library built on top of \texttt{spaCy} \citep{scispacy}. It contains neural entity extraction trained on biomedical corpora using a bidirectional-LSTM with a conditional random field (CRF) layer as proposed in \cite{lample-etal-2016-neural}. \texttt{scispaCy} identifies clinical and biomedical terms on the text first with its entity recognition model, and then retroactively maps this to UMLS using a string match over synonyms. 
    \item BERT-based clinical entity extraction models such as \cite{alimova_bert_ner}, which combine a transformer architecture with CRFs and other layers that are good at entity identification. These models are considered state-of-the-art in neural entity extraction, but are fairly slow and cannot easily run on our servers, which we discuss below. While we cannot easily compare to \cite{alimova_bert_ner} due to the lack of labelled data to train the deep model, we measure latency of running BERT on a sequence of clinical notes as a proof-of-concept. We use DistilBERT as our base BERT model because of its compactness \citep{Sanh2019DistilBERTAD}, and train on a custom vocabulary which is smaller than that of the original BERT model \citep{devlin-etal-2019-bert}.
\end{itemize}
While it is difficult to quantitatively compare these methods because we lack gold-standard entity labels for our dataset, we find that the trie-based method is significantly faster than our three other comparisons with little to no loss in recognition quality. 

Note that all of the learned models also preclude us from making easy changes to our ontology-- it is difficult to retrain these models without sufficient labelled data of a given clinical concept, which may not exist. On the other hand, our trie-based approach is reasonably fast and trivial to extend. We find that it is suitable for our purposes. 

\begin{table}[]
    \centering
    \begin{tabular}{ccp{0.5\linewidth}}
         System & Latency (seconds) & Comments \\
         \midrule
         Trie-based & 0.8 & Ours, poor disambiguation for the few overloaded concepts \\ 
         cTakes & 37 & Provides virtually the same extraction as the trie-based procedure, but with certainty/polarity scores \\ 
         scispaCy & 19.5 & Bulk of the time spent on mapping extracted terms to UMLS. Some acronyms were not diambiguated, e.g. \texttt{dm} was extracted as both \texttt{diabetes mellitus} and \texttt{double miutes}  \\
         DistilBERT & 489 & No extraction, just passing windowed snippets of the text through a compact transformer\\ 
    \end{tabular}
    \caption[Comparing NER approaches on OMR notes]{Comparing NER approaches on OMR notes both by latency and by qualitative ability to extract concepts well. Latency is measured by time to process 100 randomly chosen OMR notes.}
    \label{ner_omr_notes}
\end{table}

\subsection{Bucketization of Triage Vitals}\label{triage_vital_appendix}
As described in Section \ref{symptom_autocomplete}, our best model for predicting a ranked list of relevant symptoms to document relied on a categorical featurization of triage vitals. The model simply uses the empiric frequencies of symptoms documented in a note, conditioned on the chief complaint $c$ and a categorical representation $b(v)$ of the most abnormal vital $v$. We used medical guidelines to determine cutoffs for each vital as follows:
\begin{itemize}
    \item \emph{Temperature:} Temperatures above 100.4$^{\circ}$ are considered \texttt{HIGH} as they are medical-grade fevers. Temperatures below 97$^{\circ}$ are considered \texttt{LOW} as they are hypothermic. Otherwise, a temperature is considered \texttt{NORMAL}. 
    \item \emph{Respiratory rate:} A respiratory rate above 20 breaths per minute is considered \texttt{HIGH}, as per \citep{resp_rate}. A respiratory rate below 12 breaths per minute is considered \texttt{LOW}. Otherwise, the respiratory rate is considered \texttt{NORMAL}. 
    \item \emph{Blood oxygen level}: A pulse oximeter reading below 95\% is considered \texttt{LOW} as per Mayo Clinic guidelines. Otherwise, the reading is considered \texttt{NORMAL}. 
    \item \emph{Heart rate:} A heart rate above 100 beats per minute (bpm) is considered \texttt{TACHYCARDIC}. A heart rate below 60 is considered \texttt{BRADYCARDIC}. Otherwise, it is considered \texttt{NORMAL}. 
    \item \emph{Blood pressure:} Based on guidelines set by the American Heart Association, a systolic BP under 120 mmHg and a diastolic BP under 80 mmHg constitutes a \texttt{NORMAL} BP. If the diastolic BP is under 80 mmHg but the systolic BP is between 120-130 mmHg, it is considered \texttt{ELEVATED} blood pressure. If the systolic BP is under 140 mmHg and the diastolic blood pressure is under 90 mmHg, this is characterized as \texttt{STAGE 1 HYPERTENSION}. Otherwise, if either reading is higher, it is \texttt{STAGE 2 HYPERTENSION}. 
    \item \emph{Age:} Based on the age distribution of patients in the hospital, we bucketized patients into six groups: \texttt{CHILD} (e.g. below 18), \texttt{18-33}, \texttt{34-48}, \texttt{48-64}, \texttt{64-77}, and \texttt{78+}. 
\end{itemize}
    
\section{Extended Autocomplete Performance}
\subsection{Autocompletion Scope and Type Detection}\label{scopetype_appendix}
Here, we provide an algorithm sketch of our autocompletion scope and type detection framework. The algorithm greedily uses keywords that act as autocompletion triggers, and is run and updated as a physician types a clinical note. 
First, we initialize the scope and type of our autocompletion to be null. Then, for each word $w$ in the text, we update the scope accordingly:
\begin{itemize}
    \item If $w$ is part of a autocompletion trigger phrase such as \texttt{presents with}, we turn the autocompletion scope on and suggest terms to the user. We set the autocompletion type based on the trigger (\texttt{presents with} maps to \texttt{SYMPTOM}.)
    \item If $w$ is a continuation token such as \texttt{and}, \texttt{or}, or \texttt{,}, we maintain the current scope and autocompletion type. 
    \item If $w$ is part of a tagged concept $c$, we turn the autocompletion scope on, and set the autocompletion type to the concept type of $c$. 
    \item Otherwise, $w$ is treated as a stopword, in which case the autocompletion scope is turned off. 
\end{itemize}
With this framework, the autocompletion scope and type is greedily set using a simple parsing algorithm that is rerun as the user types a new word. 

\begin{figure}
    \centering
    \subfigure[][Manual autocompletion trigger]{\includegraphics[width=0.4\textwidth]{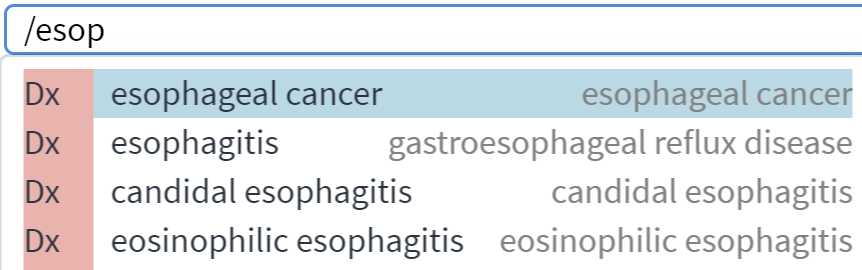}}
    \hspace{1cm}
   \subfigure[][Retroactive tagging]{\includegraphics[width=.3\textwidth]{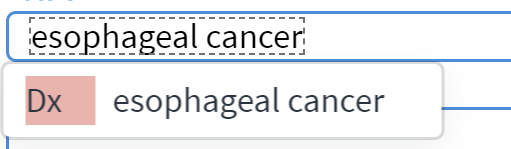}}
    \caption{Screenshots of our backup data capture strategies in the case that autocompletion scope detection algorithms fail. (a) Users can insert a slash character (\texttt{/}), which acts as a manual trigger to force autocompletion. (b) Users can retroactively accept tags for candidate concepts that they typed but did not autocomplete.}
    \label{retrospective_annotation_sc}
\end{figure}

\subsection{Training Contextual Model for Conditions}\label{training_appendix}
Our contextual model to predict a ranking over conditions is a dual-branched network that takes in two inputs:
\begin{enumerate}
    \item A Term Frequency-Inverse Document Frequency (TF-IDF) representation of the triage text using unigrams and bigrams. Vocabulary size is close to 22,000. 
    \item The binary presence of different \emph{model relevancy buckets} (as defined in Section \ref{dataset_labels}) in the patient's prior medical history. This is a length-227 binary vector.
\end{enumerate}
These two inputs are both passed through two separate dense layers with ReLU activation, concatenated and passed through another dense layer, and then finally passed through element-wise sigmoid activations to generate probabilities per class. We train this model with stochastic gradient descent using a cross entropy loss function.

\subsection{Retrospective Autocompletion Performance using MRR, MAP, and Keystroke Burden}\label{retrospective_performance_appendix}
From an information retrieval perspective, we can analyze the quality of our ranked list of suggested clinical concepts by using two standard metrics: the \emph{mean reciprocal rank} (MRR) and \emph{mean average precision} (MAP). Consider an ordered ranking $R = \{ r_1, r_2, r_3, \cdots \}$ of suggested terms and a ground truth set of terms that the clinician wants to documented denoted by $T = \{ r_{\pi(1)}, r_{\pi(2)}, r_{\pi(3)}, \cdots \}$. We define the MRR of these suggestions as $$MRR = \frac{1}{|T|}\sum_{\{ r_i \in R | r_i \in T\}} (\max(1, i - |T|))^{-1}$$ In other words, this measures the average excess rank of the suggested terms that actually occur in the ground-truth terms the clinician wants to document. An MRR of 1 indicates that $k$ desired terms were in the top $k$ suggestions. The MAP score, in contrast, measures the average proportion of ground-truth terms that occur in the top $k$ suggested terms as $k$ varies: $$MAP = \frac{1}{|T|} \sum_{k=1}^{|T|} \text{AveP}(k)$$ where $\text{AveP}(k)$ represents average precision of the top $k$ suggested terms. A MAP of 1 indicates perfect precision.

Below, we compare the various models we prototyped to predict each clinical concept type with MAP and keystroke burden. Results in terms of MRR are in Figure \ref{retrospective_evaluation}.

\subsection{Ontologies and Code}\label{ontologies_appendix}
The codebase for our analyses as well as our publicly-available ontologies for conditions, symptoms, labs, and medications can be found here: \url{https://github.com/clinicalml/ContextualAutocomplete_MLHC2020}.

\begin{figure}
    \centering
    \begin{tabular}{p{9cm}|p{2cm}|p{2cm}}
    \toprule 
    Model Type & Keystroke Burden $\downarrow$ & MAP $\uparrow$ \\
    \midrule 
     \textbf{Conditions} & & \\ 
     Frequency-based baseline & 3.44 {\tiny $\pm 0.09$} & 0.08 {\tiny $\pm 0.01$}\\
     One vs. Rest Logistic Regression on triage text $T$ & 3.02 {\tiny $\pm 0.09$} & 0.08 {\tiny $\pm 0.02$}\\ 
     OvR LR on $T$, EHR & 2.81 {\tiny $\pm 0.08$} & 0.15 {\tiny $\pm 0.02$}\\
     Augmented OvR LR on $T$, EHR & 2.71 {\tiny $\pm 0.08$} & 0.16 {\tiny $\pm 0.01$}\\
     Dual-branched neural network & 2.57 {\tiny $\pm 0.07$}& 0.27 {\tiny $\pm 0.02$}\\ 
     \midrule 
     \textbf{Symptoms} &  & \\ 
     Empirical Conditioning on Chief Complaint &  2.19{\tiny $\pm 0.04$} & 0.41 {\tiny $\pm 0.01$} \\
     Empirical Conditioning on Chief Complaint, Vital & \textbf{2.09} {\tiny $\pm 0.03$} & 0.44 {\tiny $\pm 0.01$}\\
     One vs. Rest Logistic Regression & 2.74{\tiny $\pm 0.02$} & 0.16 {\tiny $\pm 0.01$}\\ 
     One vs. Rest Naive Bayes & 2.51 {\tiny $\pm 0.03$} & 0.30 {\tiny $\pm 0.01$}\\
     \midrule
     \textbf{Labs} (ranked by frequency) &  0.092 {\tiny $\pm 0.03$} & 0.39 {\tiny $\pm 0.01$}\\
     \midrule
     \textbf{Medications} (ranked by frequency) & 3.28 {\tiny $\pm 0.04$}  & 0.03 {\tiny $\pm 0.01$}\\
     \midrule
     \textbf{Overall} with autocomplete scope/type detection & 3.13 {\tiny $\pm 0.05$} & 0.27 {\tiny $\pm 0.06$} \\
     \bottomrule
    \end{tabular}
    \caption{Retrospective Evaluation of Keystroke Burden and MAP using Contextual Autocompletion. We report the mean keystroke burden/MAP for the contextual autocomplete models we prototyped for each concept type, following the conventions of Figure \ref{retrospective_evaluation}}
\end{figure}

\begin{figure}
    \centering
    \includegraphics[height=10cm]{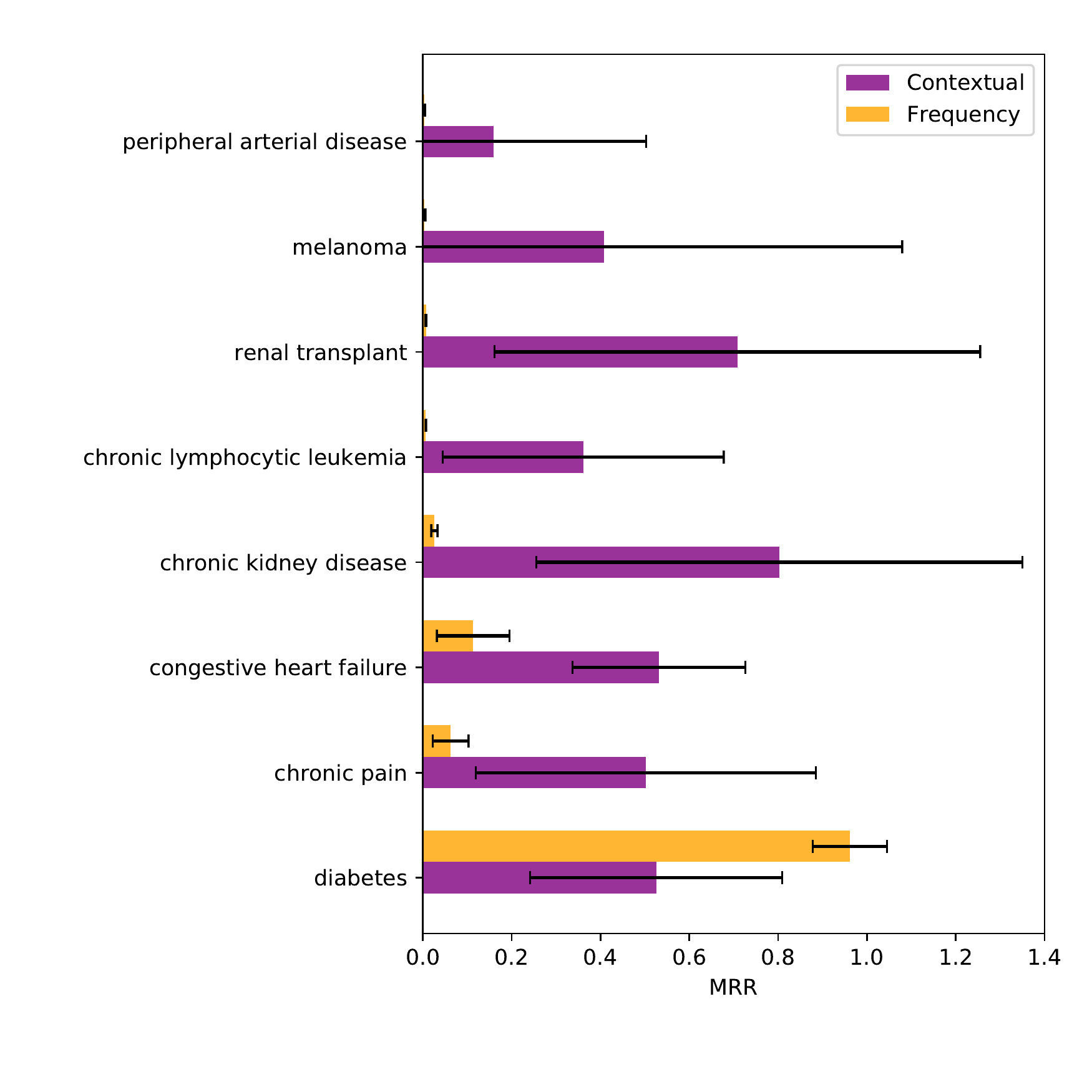}
    \caption{Mean MRR for five conditions ($\pm$ 95\% CI from mean) using contextual and frequency-based autocompletion. Concepts were chosen to get representative samples of the data. }
    \label{mrr_by_concept}
\end{figure} 

\end{document}